\pdfoutput=1

\documentclass[11pt]{article}

\usepackage{acl}
\usepackage{comment}

\usepackage{times}
\usepackage{latexsym}

\usepackage[T1]{fontenc}
\usepackage[utf8]{inputenc}

\usepackage{cancel}
\usepackage{microtype}
\usepackage{amsfonts}
\usepackage{amsmath}
\usepackage{amssymb,bm}
\usepackage{mathtools}
\usepackage{amsthm}
\usepackage{bbm}
\usepackage{scalerel}
\usepackage{xspace}
\usepackage{subcaption}
\usepackage{multirow}
\usepackage{enumitem}
\usepackage{adjustbox}
\usepackage{booktabs}
\usepackage{xfrac}

\usepackage[capitalize,noabbrev]{cleveref}
\crefname{section}{\S}{\S\S}
\crefname{table}{Tab.}{Tab.}
\crefname{figure}{Fig.}{Figs.}
\crefname{algorithm}{Alg.}{}
\crefname{equation}{Eq.}{Eq.}
\crefname{appendix}{App.}{}
\crefname{theorem}{Theorem}{}
\crefname{prop}{Proposition}{}
\crefname{cor}{Corollary}{}
\crefname{observation}{Observation}{}
\crefname{assumption}{Assumption}{}
\crefname{hyp}{Hyp.}{Hypotheses}
\crefformat{section}{\S#2#1#3}
\crefname{namedtheorem}{Hyp.}{Hypotheses}

\newtheorem{theorem}{Theorem}[section]

\newtheorem{myexample}[theorem]{Example}

\newcommand{\suggests}[2]{{#2}}

\definecolor{MacroColor}{RGB}{0,120,148}
\newcommand{\mymacro}[1]{#1}
\newcommand{\defeq}[0]{\mathrel{\stackrel{\textnormal{\tiny def}}{=}}}
\newcommand{\yy}{\mymacro{\boldsymbol{y}}}

\newcommand{\vocab}{\mymacro{\mathcal{V}}}
\newcommand{\eos}{\mymacro{\textsc{eos}}\xspace}

\newcommand{\calC}{\mymacro{\mathcal{C}}}
\newcommand{\corpus}{\mymacro{\mathcal{D}}}
\newcommand{\defn}[1]{\textbf{#1}}

\newcommand{\vtheta}{\mymacro{\boldsymbol \theta}}
\newcommand{\ptheta}{\mymacro{p_{\scaleto{\vtheta}{4pt}}}}
\newcommand{\pdata}{\mymacro{p_{\scaleto{\corpus}{4pt}}}}
\newcommand{\pdec}{\mymacro{\widetilde{p}}}

\newcommand{\adapter}{\mymacro{\boldsymbol{\alpha}}}
\newcommand{\adapterdist}{\mymacro{\pdec_{\scaleto{\vtheta}{4pt}}}}
\DeclareMathOperator*{\argmax}{\mymacro{\mathrm{argmax}}}
\DeclareMathOperator*{\argmin}{\mymacro{\mathrm{argmin}}}
\newcommand{\KL}{\mymacro{\mathrm{KL}}}
\newcommand{\JS}{\mymacro{\mathrm{JS}}}
\newcommand{\tvd}{\mymacro{\textsc{tvd}\xspace}}
\newcommand{\ent}{\mymacro{\mathrm{H}}}

\newcommand{\vocabeos}{\mymacro{\overline{\vocab}}}

\newcommand{\mauve}{\mymacro{\textsc{Mauve}\xspace}}
\newcommand{\threeS}{\mymacro{{\tiny\ensuremath{{}^{\textstyle *}}}}}

\newcommand{\gptj}{\mymacro{GPT-J\xspace}}

\newcommand{\langmodel}{\mymacro{\ptheta(\cdot \mid\yy_{<t})}}

\usepackage{todonotes}
\makeatletter
\newcommand*\iftodonotes{\if@todonotes@disabled\expandafter\@secondoftwo\else\expandafter\@firstoftwo\fi}  %
\makeatother

\title{On the Efficacy of Sampling Adapters}

\usepackage{inconsolata}

\usepackage{emoji}

\newcommand{\ucambridge}{\emoji[emoji]{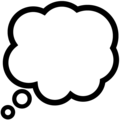}}
\newcommand{\ethz}{\emoji[emoji]{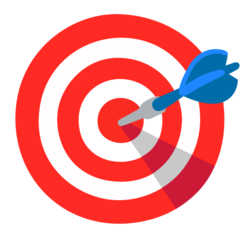}} 

\author{ Clara Meister$^{\ethz}$~\;~Tiago Pimentel$^{\ucambridge}$~\;~ Luca Malagutti$^{\ethz}$\\
\bf{Ethan G. Wilcox}$^{\ethz}$~\;~\bf{Ryan Cotterell}$^{\ethz}$ \\
   $^{\ethz}$ETH Zürich~\;~ $^{\ucambridge}$University of Cambridge\\
  \texttt{\href{mailto:meistecl@inf.ethz.ch}{meistecl@inf.ethz.ch}}~\;~\texttt{\href{mailto:tp472@cam.ac.uk}{tp472@cam.ac.uk}}~\;~\texttt{\href{mailto:lmalagutti@inf.ethz.ch}{lmalagutti@inf.ethz.ch}}\\
   \texttt{\href{mailto:ethan.wilcox@inf.ethz.ch}{ethan.wilcox@inf.ethz.ch}}~\;~\texttt{\href{mailto:ryan.cotterell@inf.ethz.ch}{ryan.cotterell@inf.ethz.ch}}
   }

\begin{document}
\maketitle
\begin{abstract}
Sampling is a common strategy for generating text from probabilistic models, yet standard ancestral sampling often results in text that is incoherent or ungrammatical. 
To alleviate this issue, various modifications to a model's sampling distribution, such as nucleus or top-$k$ sampling, have been introduced and are now ubiquitously used in language generation systems.
We propose a unified framework for understanding these techniques, which we term \defn{sampling adapters}.
Sampling adapters often lead to qualitatively better text, which raises the question: From a formal perspective, how are they changing the (sub)word-level distributions of language generation models? And why do these local changes lead to higher-quality text? 
We argue that the shift they enforce can be viewed as a trade-off between precision and recall: while the model loses its ability to produce certain strings, its precision rate on desirable text increases. 
While this trade-off is not reflected in standard metrics of distribution quality (such as perplexity), we find that several precision-emphasizing measures indeed indicate that sampling adapters can lead to probability distributions more aligned with the true distribution. 
Further, these measures correlate with higher sequence-level quality scores, specifically, \mauve.\looseness=-1 

\vspace{1.0em}
    {\includegraphics[width=1.25em,height=1.0em]{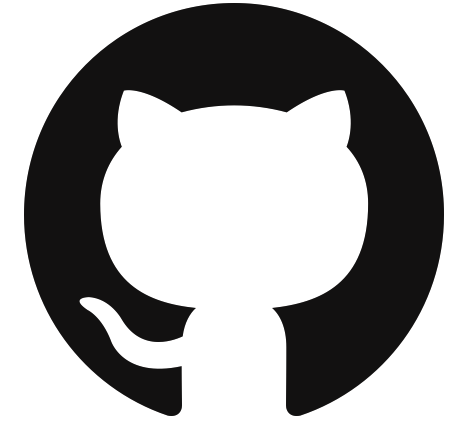}\hspace{1em}\parbox{\dimexpr\linewidth-2\fboxsep-2\fboxrule}{\url{https://github.com/rycolab/sampling-adapters}}}
\vspace{0.5em}
\end{abstract}

\section{Introduction}

The vast majority of natural language generation systems take a probabilistic approach. %
The backbone of such an approach is a probability distribution over strings $\ptheta$ for a specific target domain. 
While modern language models have achieved remarkable performance on standard measures of distribution quality, e.g., perplexity \cite{NEURIPS2020_GPT3,PALM,hoffmann2022an,openai2023gpt4}, they often fall short when applied out of the box for language generation tasks---both sampling directly from them and searching for the maximum-probability string under them can lead to dull, incoherent, and degenerate text \cite{holtzman_curious_2020,eikema_is_2020,welleck2019neural}.\looseness-1 

Surprisingly, applying a post-hoc modification to $\langmodel$ often serves to dramatically improve the quality of the generated text \cite{nadeem-2020-systematic,pillutla2021mauve,wiher+al.tacl22,hewitt2022truncation,li2022contrastive}.
In this paper, we give a name to these methods, dubbing them \defn{sampling adapters}. 
A sampling adapter can be formally defined as a simplex-to-simplex map
$\adapter\!: \Delta^{|\vocabeos|-1} \rightarrow \Delta^{|\vocabeos|-1}$ that systematically modifies the conditional distribution of an autoregressive language model $\langmodel$, thus creating another language model $\adapter(\langmodel)$ with a desired set of characteristics, e.g., it may only give non-zero probability to items assigned high probability under the original model.
Sampling adapters often require little to no fine-tuning and can be implemented in just a few lines of code. 
Presumably due to their simplicity, sampling adapters have become a default tool in text generation pipelines, serving as the core component of baseline decoding strategies in various tasks \cite{welleck2019neural,pillutla2021mauve,pimentel2023on}.\looseness=-1

The fact that sampling adapters often lead to qualitatively better text, however, evokes a simple question: How do they change our language generation models such that the distribution $\langmodel$ places more probability mass on what we qualitatively deem to be ``better'' text? 
Most sampling adapters have been found through trial and error with only intuitive motivations given for their efficacy. 
Moreover, standard evaluation measures\footnote{We use the term \emph{measure} instead of the more common \emph{metric} throughout this work because several of the functions that we consider are not metrics in the mathematical sense.} do not immediately shed light on why sampling adapters work well because most sampling adapters make language generation models substantially worse according to these measures, e.g., they often reduce the probability assigned to certain strings to zero, which can yield a perplexity of $\infty$.\looseness=-1 

In this paper, we posit that the change of distribution induced by sampling adapters can be analyzed in terms of a precision--recall trade-off, using the generalizations of these terms to the field of generative modeling \cite{sajjadi2018assessing, lucic2018gans, djolonga2020precision}. 
While a model loses its ability to produce certain strings, its ability to produce \emph{desirable} text increases. 
We experiment with various sampling adapters that have been proposed \cite{fan_hierarchical_2018,holtzman_curious_2020,meister+al.pre22,hewitt2022truncation} and find that, while the use of these adapters negatively affects recall-emphasizing performance measures, certain choices of hyperparameters increase performance in terms of measures that balance between precision and recall or that are precision-emphasizing.  
Comparing trends in these measures, we see evidence of a precision--recall trade-off, which offers a quantitative motivation for the efficacy of sampling adapters. 
We further find that precision-emphasizing measures correlate most highly with sequence-level quality metrics, offering a potential avenue for efficiently choosing sampling adapter hyperparameter values. 
The formal framework and empirical analysis presented here should pave the way for the development of theoretically motivated sampling adapters, and provide a straightforward means for both analysis of and comparison between adapters.\looseness=-1

\section{Language Generation}

\subsection{Probability Distributions over Strings}
Most language generation systems are based on probabilistic models, i.e., models of the probability distribution over natural language strings\footnote{Notably, these distributions might be conditioned on an input string, as in machine translation or summarization.\looseness=-1} $\vocab^*$, 
where $\vocab^*$ is the Kleene closure of an alphabet $\vocab$.
In words, $\vocab^*$ is the set of all strings that can be generated from a vocabulary of (sub)words $\vocab$.
A common modeling choice is to break down string probabilities autoregressively and locally normalize $\ptheta$, i.e., instead of directly modeling the full sequence probability $\ptheta(\yy)$, one models (sub)word probabilities $\ptheta(y \mid \yy_{<t})$ conditioned on the prior context $\yy_{<t}\defeq\langle y_1, \dots, y_{t-1}\rangle \in \vocab^*$. 
Note that here, we have $y\in \vocabeos$ for $\vocabeos \defeq \vocab\cup \{\eos\}$ where \eos is a special \underline{e}nd \underline{o}f \underline{s}tring token required for an autoregressive $\ptheta$ to define a valid probability distribution over $\vocab^*$. 
The sequence-level probability can then be computed via the chain rule of probability:\looseness=-1
\begin{equation}
    \ptheta(\yy) = \ptheta(\eos \mid \yy) \prod_{t=1}^{|\yy|} \ptheta(y_t \mid \yy_{<t})
\end{equation}
\noindent
See \citet{du2022measure} for a characterization of when these models are tight, i.e., when the probability mass assigned to finite-length strings is 1.

The parameters $\vtheta$ of these models are typically chosen by (numerically) maximizing the log-likelihood of the training data $\corpus$, where log-likelihood is defined as:\looseness=-1
\begin{equation}\label{eq:likelihood}
\mathcal{L}(\vtheta) \!=\! \sum_{\yy\in \corpus}\log \ptheta(\yy) 
\end{equation}
Note this is equivalent to minimizing the (forward) cross-entropy between the empirical distribution $\pdata$ induced by the training data $\corpus$.

\subsection{Decoding Strategies}
In order to produce text from a model, one must use a \defn{decoding strategy}, which provides a set of decision rules according to which tokens are sequentially chosen from the distribution $\ptheta$ to form a string. 
Decoding strategies can be broadly taxonomized as either maximization-based or sampling-based.
Maximization-based strategies aim to find the candidate string that scores highest under some objective.
Finding the string with the highest probability under the model is a common maximization-based strategy.
Sampling-based strategies instead \emph{sample} tokens according to some distribution derived from the model.
While maximization-based strategies may make intuitive sense, they often lead to dull or degenerate text in open-generation settings \cite{pmlr-v97-cohen19a,eikema_is_2020,nadeem-2020-systematic}.
Sampling-based strategies likewise have shortcomings: They introduce randomness into the generated text, which may lead to a disruption in coherence or fluency when units are sampled from low-probability regions of the distribution \cite{holtzman_curious_2020,hewitt2022truncation}. 
A class of methods has been developed to address the problems observed when sampling directly from the model, specifically by altering the distribution from which tokens are sampled.   
We term these methods sampling adapters, formally defining them in the next section.\looseness=-1

\section{The Sampling Adapter Framework}
Formally, sampling adapters are simplex-to-simplex mappings, i.e., functions $\adapter\!: \Delta^{|\vocabeos|-1} \rightarrow \Delta^{|\vocabeos|-1}$ that take a probability distribution over $\vocabeos$ as input and map it to another one over $\vocabeos$.\footnote{Sampling adapters can be generalized to work on full distributions $p(\yy)$ instead of on the conditionals $p(\cdot \mid \yy_{<t})$, but we focus on the simpler case of the conditionals here.}
We use the notation $\pdec$ to denote the output of this map, as applied to the distribution $p$:
\begin{align}
\pdec(\cdot \mid \yy_{<t}) \defeq \adapter\big(p(\cdot \mid \yy_{<t})\big) \label{eq:adapter}
\end{align}
similarly denoting the individual adapted probabilities as $\pdec(y \mid \yy_{<t}) = \adapter\big(p(\cdot \mid \yy_{<t})\big)(y)$.
We now give two examples of common sampling adapters. 
\begin{myexample}
We recover standard \defn{ancestral sampling} when
$\adapter\big(p(\cdot \mid \yy_{< t})\big)(y)= p(y \mid \yy_{<t})$.
\end{myexample}
\begin{myexample}
We recover \defn{temperature sampling} when
$\adapter\big(p(\cdot \mid \yy_{< t})\big)(y) \propto p(y \mid \yy_{<t})^{\frac{1}{T}}$ for temperature parameter $T$.\footnote{$T$ allows us to control the entropy of the distribution. As $T \rightarrow 0$, we recover a distribution that places probability 1 on the $\mathrm{argmax}$, and, as $T \rightarrow \infty$, we recover the uniform distribution.\looseness=-1}
\end{myexample}
One popular way of formulating sampling adapters in the literature has been via truncation functions, i.e., functions where vocabulary units that do not meet a certain criterion are re-assigned zero probability. We write these functions as:
\begin{align}\label{eq:truncation}
   \adapter\big( p&(\cdot \mid \yy_{< t})\big)(y) \propto \\
   &\qquad\quad p(y \mid \yy_{<t}) \mathbbm{1}\Big\{ y \in \calC\big(p(\cdot \mid \yy_{< t})\big) \Big\} \nonumber
\end{align}
where $\calC: \Delta^{|\vocabeos|-1 } \rightarrow \mathcal{P}(\vocabeos)$ is a function that finds the set of (sub)words that meets said criterion; $\mathcal{P}(\cdot)$ denotes the powerset operator.
Truncation sampling methods aim to eliminate probability mass placed on tokens deemed likely to lead to undesirable text, reallocating their probability mass to the remaining options. 
We now specify several common truncation-based sampling adapters.\looseness=-1 
\begin{myexample}
We recover \defn{top-$k$ sampling} \cite{fan_hierarchical_2018} when 
\begin{align}
    \calC(p(\cdot \mid \yy_{<t})) = &\argmax_{\vocab' \subseteq \vocabeos} \sum_{y \in \vocab'} p(y \mid \yy_{<t})\\
    &\,\,\text{s.t.}\,\,|\vocab'|=k\nonumber
\end{align}
i.e., a function that returns the top-$k$ most-probable (sub)words.\looseness=-1
\end{myexample}

\begin{myexample}
We recover \defn{top-$\pi$ (nucleus) sampling} \cite{holtzman_curious_2020}  when  
\begin{align}
    \calC(p(\cdot \mid \yy_{<t})) = &\argmin_{\vocab' \subseteq \vocabeos} |\vocab'|\\
    &\,\,\text{s.t.}\,\,\sum_{y \in \vocab'} p(y \mid \yy_{<t}) \geq \pi \nonumber
\end{align}
i.e., a function that returns the smallest subset of (sub)words that collectively have probability mass $\geq\pi$.\looseness=-1
\end{myexample}
\begin{myexample}
We recover \defn{locally typical sampling} \cite{meister+al.pre22} when
\begin{align}
    \calC(p(\cdot \mid \yy_{<t})) &= \argmin_{\vocab' \subseteq \vocabeos} \sum_{y\in\vocab'} \Big|\ent(p(\cdot \mid \yy_{<t}))\\
    &\qquad\qquad\qquad\quad\,\,+\log p(y \mid \yy_{<t})\Big| \nonumber\\
    &\,\,\text{s.t.}\,\,\sum_{y \in \vocab'} p(y \mid \yy_{<t}) \geq \pi \nonumber
\end{align} i.e., the set of items with log-probability closest to the (sub)word-level entropy that collectively have probability mass $\geq\pi$.
\end{myexample}
\begin{myexample}
We  recover \defn{$\eta$-sampling}~\cite{hewitt2022truncation} when 
\begin{align}
    \calC(p(\cdot \mid \yy_{<t})) = \{y \in \vocabeos \mid p(y \mid \yy_{<t}) > \eta\}
\end{align}
where $\eta = \min 
\left(\epsilon, \sqrt{\epsilon}\exp(-\ent\left(p(\cdot \mid \yy_{<t})\right))\right)$, i.e., the set of items with probability greater than $\eta$ for hyperparameter $\epsilon > 0$.\looseness=-1
\end{myexample}
Other methods can similarly be cast in the sampling adapter framework, such as Mirostat \cite{BasuRKV2021} and the re-calibration method proposed by \citet{pmlr-v119-braverman20a}. 
Moreover, the general equation for sampling adapters given in \cref{eq:adapter} suggests that one direction for future research is \emph{learning} a sampling adapter $\adapter$. 
While many previously proposed adapters are truncation-based, 
adapters that reallocate mass in a different manner may also prove effective. Indeed, equipping $\adapter$ with tunable parameters could prove useful as a lightweight fine-tuning method.

\paragraph{An Unintuitive Effect.}
The motivation behind the use of sampling adapters with language generation models is to readjust their distribution, shifting mass away from tokens deemed likely to lead to undesirable text and onto tokens that will generate high-quality text. 
Yet why are such transformations even necessary? 
Standard measures of distribution quality, such as perplexity, would suggest that our models' estimates of the ground-truth distribution over natural language strings are quite good \cite{NEURIPS2020_GPT3,gpt-j,hoffmann2022an}.
This, in turn, implies that the heuristic shifts performed by sampling adapters should lead to \emph{worse} language generators. We argue that the disparity between the quality of language generation systems using sampling-adapted models and the quality of these same models according to standard measures can be reconciled using probabilistic analogs of precision and recall.\looseness=-1

\section{A Precision--Recall Hypothesis}

We begin by reviewing generalizations of the concepts of precision and recall in the field of generative modeling.   
We then discuss the shortcomings of current language generation models and how sampling adapters may address these shortcomings.

\subsection{Generalizations of Precision and Recall}\label{sec:pr}
A series of recent papers have related the \defn{precision} of a learned distribution $\ptheta$ to the average quality of generated samples, where high-quality samples are assumed to be those with high probability under the data-generating distribution $p$.\footnote{We note that in general though, it is not clear that high-probability and high-quality should necessarily coincide \cite{zhang_trading_2020,meister+al.pre22}.} 
Additionally, they relate the \defn{recall} of $\ptheta$ to its coverage of $p$ \cite[][\emph{inter alia}]{sajjadi2018assessing, lucic2018gans, djolonga2020precision}, i.e., high overlap in the support of $\ptheta$ and $p$.
Following this line of reasoning, the notions of precision and recall can naturally be operationalized using measures of the difference between two distributions---specifically, ones that enable different penalizations of over- and under-coverage of our reference distribution.

There are several measures that, when considered together, naturally operationalize precision, recall, or some combination of the two.\footnote{We refer the reader to \citet{cichocki_2010_families} and \citet{djolonga2020precision} for a more comprehensive discussion of such measures.} 
In this paper, we focus on cross-entropy,  
$\KL$ divergence, total variation distance ($\tvd$), and Jensen--Shannon ($\JS$) divergence. We introduce each in greater detail below. 
We note that for all these measures, a larger value indicates a greater discrepancy between two distributions, and that all but the cross-entropy will be zero when the two distributions are identical. 
Further, we note that not all the measures are symmetric, i.e., their values change depending on the order in which the distributions are given as arguments to the measure. 
Out of convention, in the case that the reference distribution is provided first, we call this the \defn{forward} variant of the measure. 
We call the case where the reference distribution is the second argument the \defn{reverse} variant of the measure.
We define all measures in terms of generic distributions $p_1$ and $p_2$, which we assume both have (not necessarily identical) supports that are a subset of $\vocabeos$.\looseness=-1

\paragraph{Precision-emphasizing Measures.}
We first consider the \defn{cross-entropy} between $p_1$ and $p_2$:\looseness=-1
\begin{align}\label{eq:xent}
        \ent(p_1, p_2) = -\sum_{y\in \vocabeos} p_1(y) \log p_2(y) 
\end{align}
Upon inspection, we can see that the reverse cross-entropy, i.e., where $p_1$ is the distribution being evaluated and $p_2$ is a (fixed) reference distribution, rewards high precision.\footnote{We note that most readers are likely more familiar with the \emph{forward} cross-entropy, which is a common loss function.}   
Specifically, it rewards $p_1$ for assigning probability mass where $p_2$ is large, implicitly penalizing $p_1$ for assigning high probability where $p_2$ is small.  
In fact, the reverse cross-entropy is minimized in the case where $p_1$ places all probability on the most probable token under $p_2$.

A related measure is the reverse $\KL$ divergence
\begin{subequations}
\begin{align}\label{eq:KL}
        \KL(p_1 \mid\mid p_2) &=  \sum_{y\in\vocabeos}p_1(y) \log\frac{p_2(y)}{p_1(y)}\\
         &= \ent(p_1, p_2) - \ent(p_1)
\end{align}
\end{subequations}
which is equivalent to the cross-entropy up to the subtraction of the entropy term $\ent(p_1)$.  
As with cross-entropy, the reverse $\KL$ divergence rewards high precision. 
This property is reflected by a common intuition provided about this measure when it is used as a learning objective: It is referred to as a \emph{mode-seeking} objective, i.e., it aims to place mass on the \emph{modes} of $p_1$.\footnote{For further insights about the properties of the various measures used here, we refer the reader to the following detailed discussions  \citep{minka_divergence,JMLR:v9:nickisch08a,huszar_how_2015,theis_note_2016}.\looseness=-1}
Importantly, the distributions that minimize the reverse variants of \cref{eq:xent,eq:KL} will not necessarily be equivalent because the latter takes into account $p_1$'s entropy. 
So which of these two metrics should we use? As we are interested in using metrics that operationalize the notion of precision, the entropy of the distribution under evaluation is irrelevant. Thus, we will use the reverse cross-entropy as our primary precision-emphasizing metric.

\paragraph{Recall-emphasizing Measures.}
On the other hand, the forward variants of \cref{eq:xent,eq:KL}, where $p_2$ is now the distribution under evaluation and $p_1$ is assumed to be fixed, reward recall.
This is evident when taking a closer look at their definitions.
If $p_2$ fails to place probability on all elements $y$ assigned probability by $p_1$, then both the cross-entropy and $\KL$ divergence will be $\infty$.\footnote{To avoid the possibility of an infinite cross-entropy, one can use an $\varepsilon$-smoothed variant of $p_2$ i.e., where $p_2^{(\varepsilon)}(\cdot) = \frac{p_2(\cdot) + \varepsilon}{1 + |\vocabeos| \cdot \varepsilon}$. 
This trick is often employed to evaluate methods that do not produce distributions covering the entire support, e.g., \citet{peters-etal-2019-sparse} and \citet{martins-etal-2020-sparse}. As many of the sampling adapters that we analyze produce sparse distributions (specifically, the truncation sampling methods), we will likewise employ this variant of $\KL$ divergence where necessary.\looseness=-1} 
Analogously to the reverse $\KL$'s description as mode-seeking, the forward $\KL$ is referred to as \emph{mean-seeking}. 
Note that using the forward variants of cross-entropy and $\KL$ divergence as learning objectives is equivalent since $\ent(p_1)$ is constant with respect to $p_2$. 
Further, the forward $\KL$ and cross-entropy, as well as the reverse KL, are minimized when $p_2 = p_1$.\looseness=-1

\paragraph{Balanced Measures.}
The definitions for $\tvd$ and $\JS$ divergence, which are both symmetric measures, suggest a balance between the characteristics of precision and recall:
\begin{equation}
    \tvd(p_1, p_2) = \sum_{y\in\vocabeos} \left|p_1(y) - p_2(y)\right|
\end{equation}
\begin{equation}
    \JS(p_1, p_2) = \frac{\KL(p_1 \mid\mid m) + \KL(p_2 \mid\mid m)}{2}
\end{equation}
where $m(y)=\frac{p_1(y) + p_2(y)}{2}$ for $y \in \vocabeos$ is a pointwise average.  
Practically, the $\JS$ divergence can informally be viewed as an interpolation between the forward and reverse $\KL$ divergences.
Indeed, several divergences that generalize the forward and reverse $\KL$ recover the $\JS$ divergence given a particular choice of hyperparameter  \cite{huszar_how_2015,meister-etal-2020-generalized,pillutla2021mauve}. 
$\tvd$ can be similarly motivated: \citet{sajjadi2018assessing} recover $\tvd$ in their precision--recall operationalization for generative models when assigning equal importance to precision and recall. 
Further, a standard result demonstrates that the $\JS$ divergence is a lower bound on $\tvd$ \cite{61115}. With these measures in hand, we can more effectively assess the shifts to precision and recall that sampling adapters induce in a model.

\subsection{Current Modeling Shortcomings}
It is not clear that the objective with which probabilistic language generators are typically trained imparts characteristics that align with the goals of building good language generators.\footnote{Several works have explored this topic specifically, which we discuss in \cref{sec:related}.} 
Any form of maximum-likelihood training is equivalent to minimizing $\ent(\pdata, \ptheta)$---often with an additional form of regularization.
Thus, it encourages high recall: $\ptheta(y_t\mid\yy_{<t})$ must be nonzero for all tokens $y_t$ in every string $\yy$ in the training set $\corpus$ for the objective to be finite.
This, in turn, results in $\ptheta$ allocating some probability mass to all (sub)words $y\in\vocabeos$ for all contexts $\yy_{<t}$. 
In language modeling, this is perhaps a desirable property: We often care about the relative probabilities of strings, and assigning strings $0$ probability would be counter-productive towards this goal. Yet, this property can potentially prove problematic when such models are used out of the box as language generators.\footnote{
To see why, consider a simple example: A model that assigns a very small collective probability mass of $0.001$ to all (sub)words in the tail (low-probability region) of the distribution at any given generation step. 
If we sample a sequence of $200$ tokens from this (unaltered) model, there is a $1\!- \!(1\!-\!0.001)^{200}\approx20\%$ chance it will contain at least one token from the tail of the distribution, which after sampled, can have negative downstream effects, ultimately rendering the whole string incoherent \cite{holtzman_curious_2020,xia2023training}.\looseness=-1}
For language generation systems, high precision is arguably a higher priority, i.e., the goal is for all of the generated sequences to be of high quality. 
An intuitive argument for this is that a single bad output can leave a lasting poor impression on the user. 
Yet, the inability to generate a single sequence may go unnoticed---especially if the difference between that sequence and one the model can produce is a single, exchangeable token. 

In this light, a possible explanation for the efficacy of sampling adapters is as follows: While model parameters are chosen to minimize a recall-prioritizing objective, sampling adapters re-align the distribution with a more appropriate \emph{precision-prioritizing} probabilistic objective, i.e., sampling adapter hyperparameter combinations that work well perhaps do so because they minimize an objective that balances between precision and recall. 
If this is indeed the case, it should not be surprising that the transformation induced by sampling adapters leads to worse models according to standard, recall-emphasizing measures: Any generator that assigns zero probability to a valid string---as is the case when top-$\pi$ or top-$k$ sampling are applied---will have both infinite cross-entropy and perplexity with respect to the natural language distribution. 
They may, however, lead to better models according to more balanced (or even precision-emphasizing) measures, which is what we now empirically test.\looseness=-1

\section{Experiments}\label{sec:exps}
To test the hypothesis that the operations performed by sampling adapters are akin to a re-prioritization of precision over recall in the output of the model, we evaluate the effects of sampling adapters on measures that emphasize recall, precision or a balance of the two, as outlined in \cref{sec:pr}. We then observe how these measures vary as a function of the sampling adapters' hyperparameters. 
Further, we also look at these measures' Spearman correlations with \mauve, a sequence-level quality metric.\looseness=-1 

We consider five different adapters: temperature, $\eta$ (eta),  top-$\pi$, top-$k$ and locally typical sampling, each over a wide range of hyperparameters.  Note that for the latter three adapters, a smaller hyperparameter value corresponds to a larger shift between $\ptheta$ and $\adapterdist$.
For $\eta$-sampling, the reverse is true, and for temperature sampling, hyperparameter values farther from $1$ imply a larger shift. 
For reproducibility, we leverage the \href{https://huggingface.co/}{Hugging Face} framework~\cite{huggingface} and its implementation of sampling adapters for all but $\eta$-sampling, for which we rely on the original authors' implementation.\footnote{\href{https://github.com/john-hewitt/truncation-sampling}{github.com/john-hewitt/truncation-sampling}} Error bars for all plots indicate 95\% confidence intervals for the observed values; note that bars are often small enough that they are not visible.

\subsection{Setup}\label{sec:setup}
 We focus on the task of open-ended text generation.  We use GPT-2 small and large \cite{gpt-2}, as well as, GPT-Neo (small)~\cite{gpt-neo} as our generation models. 
 The main results of this paper use the test set of a public version of the WebText dataset\footnote{The dataset \suggests{may be found}{is} at \href{https://github.com/openai/gpt-2-output-dataset}{github.com/openai/gpt-2-output-dataset}.} as our reference text.  Results using the WikiText test set \cite{merity2016pointer} are qualitatively similar and can be found in \cref{app:results}.
 
\paragraph{Sequence-level Metrics.}  Following \citet{pillutla2021mauve}, we use the first $35$ tokens of samples from our reference text as a prompt to generate continuations $\yy \sim \ptheta(\cdot \mid \yy_{<t})$ until $|\yy|=512$, or \eos is sampled. We generate 1000 samples for each combination of model, sampling adapter, and hyperparameter. 
We compute \mauve scores (where higher implies the samples are closer to the reference text), aggregated over $5$ seeds, for each of these sets of text samples.
\begin{figure*}[!htb]
    \centering
\begin{subfigure}[b]{.9\textwidth}
   \includegraphics[width=\linewidth]{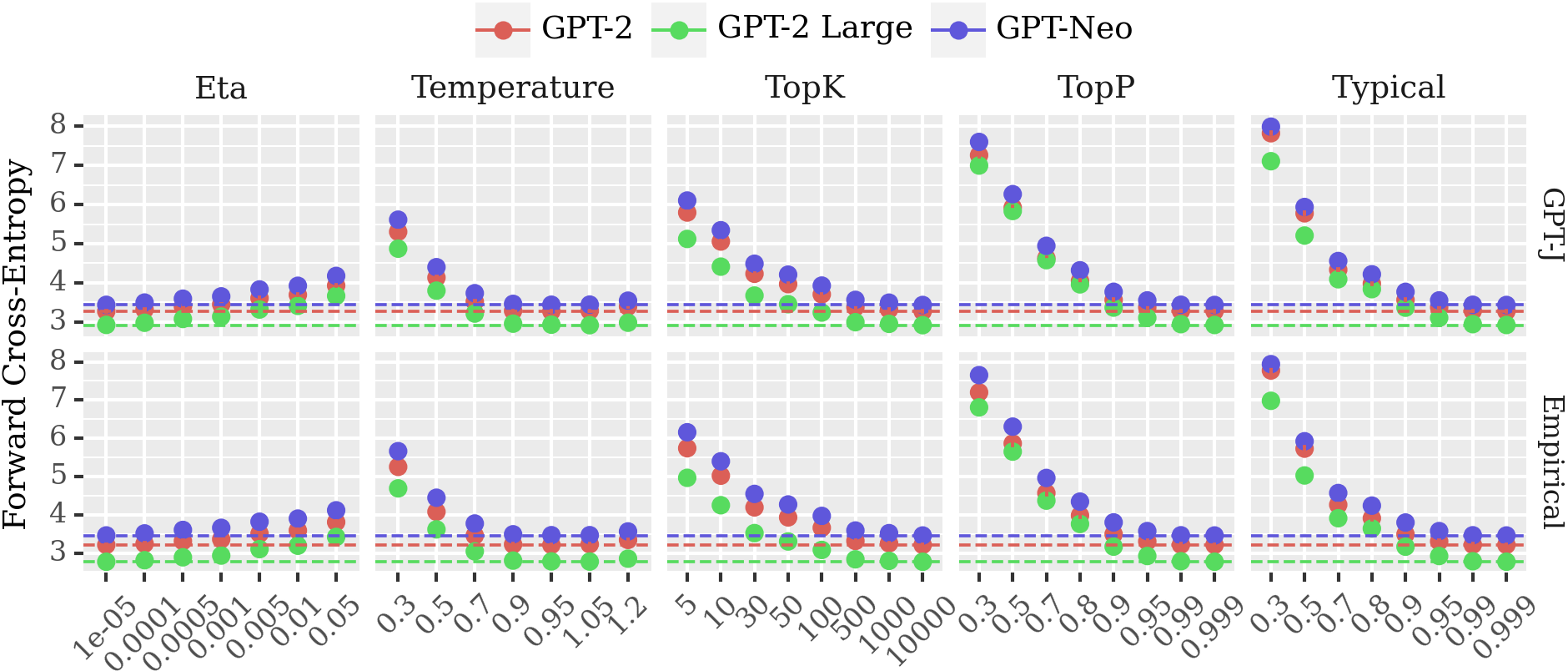}
\end{subfigure}
\begin{subfigure}[b]{.9\textwidth}
   \includegraphics[width=\linewidth]{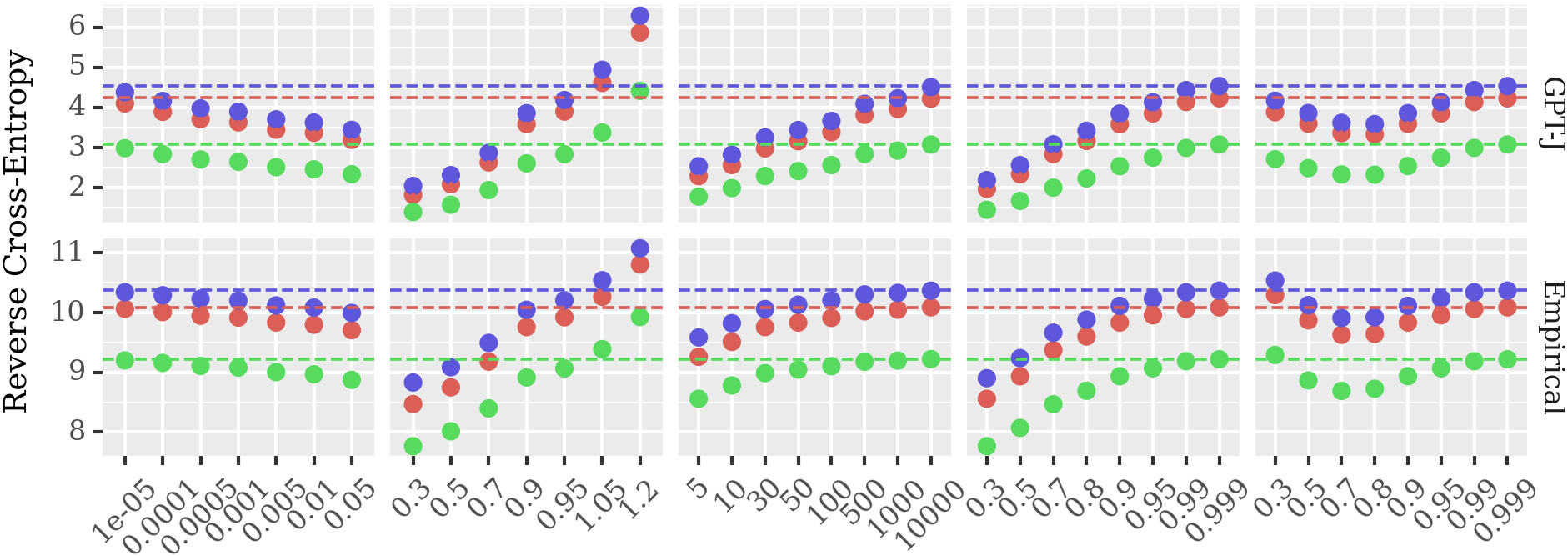}
\end{subfigure}
\begin{subfigure}[b]{.9\textwidth}
   \includegraphics[width=\linewidth]{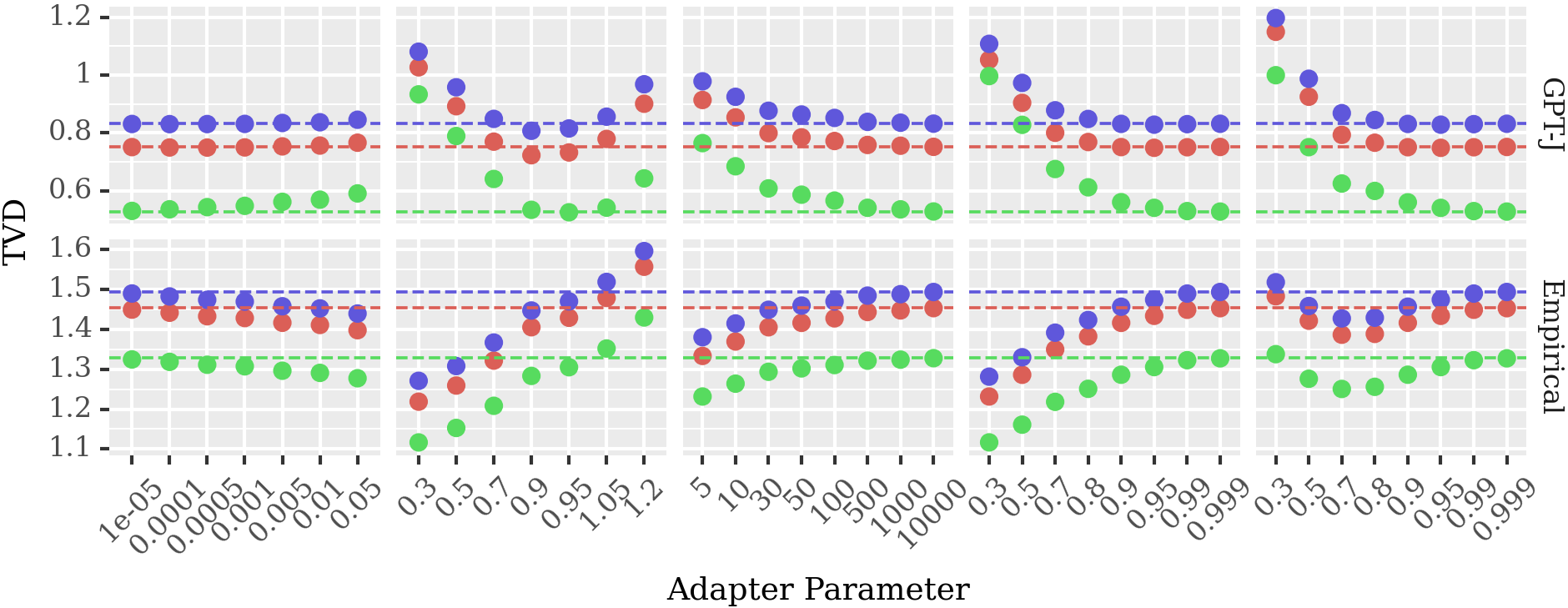}
   \vspace{-12pt}
\end{subfigure}
    \caption{Forward/reverse cross-entropy and \tvd{} of the model with GPT-J and the empirical distribution (WebText test set) after different sampling adapter methods have been applied to the output distribution. Note that as described in \cref{sec:pr}, the $\varepsilon$-variant is used in all cross-entropy estimates except for reverse estimates with GPT-J. Dashed lines represent divergence with the unmodified distribution, i.e., the equivalent of using ancestral sampling. }
    \label{fig:backward_forward}
    \vspace{-6pt}
\end{figure*}

\paragraph{Token-level Measures.}
In this analysis, we compare (sub)word-level distributions $\adapterdist(\cdot\mid\yy_{<t})$ and $p(\cdot\mid\yy_{<t})$. The former is our generation model after the application of a sampling adapter and the latter is a reference distribution.
We present results using both the empirical distribution induced by our test set and the distribution given by the GPT-J model \cite{gpt-j}\footnote{We use GPT-J as a reference because it has substantially better perplexity on benchmark datasets. Note that it has $\approx 50$ times more parameters than either GPT-2 small or GPT-Neo, both of which it shares a vocabulary with.} as our reference distribution.
Here, $\yy$ is a string from the test set. 
Results are mean-aggregated across both $t=1, \ldots, |\yy|$\ and all $\yy$. Note that when we compute either the cross-entropy or $\KL$ divergence and it is not guaranteed that the support of $p_1$ is a subset of the support of $p_2$, we make use of the $\varepsilon$ version of the metrics, as specified in \cref{sec:pr}, with $\varepsilon=1e\text{-}6$.

\subsection{Results}
\paragraph{Trends in Probabilistic Measures.} We first present our analysis of how different adapter--hyperparameter settings affect the relationship of the model to a reference distribution (either probabilities according to GPT-J or the empirical distribution).  
Note that if our hypothesis in \cref{sec:pr} is correct, we would expect to see that certain sampling adapter--hyperparameter settings lead to lower values of measures that emphasize precision, such as reverse cross-entropy, while simultaneously increasing measures that emphasize recall, such as forward cross-entropy. We show the reverse and forward cross-entropy, as well as $\tvd$, in \cref{fig:backward_forward}.\footnote{As anticipated given the relationship between $\tvd$ and $\JS$, results showing the $\JS$ divergence are qualitatively very similar to $\tvd$. Hence, they appear in \cref{app:results}. }

Both the forward and reverse cross-entropy results align closely with our hypothesis: A larger adapter shift generally leads to a higher forward cross-entropy and lower reverse cross-entropy.\footnote{Importantly, if not for use of the $\varepsilon$-smoothed versions of the forward and reverse cross-entropies,
many of the cross-entropies in \cref{fig:backward_forward} would be infinite for the truncation-based adapters. 
Specifically, this would be true for any adapter without 100\% coverage of the tokens in the evaluation text, which is the case for most adapter--hyperparameter settings (see \cref{fig:token} in \cref{app:results}).} This observation holds when using either the empirical distribution or \gptj as our reference. Interestingly, we see that the trends reverse when we consider the reverse $\KL$ divergence (as opposed to the reverse cross-entropy; see \cref{fig:backward_forward_kl}). 
This is perhaps expected given that the entropy of the model's distribution monotonically decreases after the application of sampling adapters (see \cref{fig:ent}).\looseness=-1

Lastly, the trends in $\tvd$ differ largely depending on the distribution used as a reference. When \gptj is used, we see that $\tvd$ monotonically increases as adapter strength increases. The reverse trend appears to hold when considering the empirical distribution: $\tvd$ generally \emph{decreases} with adapter strength. The reason for this difference is not immediately obvious. Closer inspection reveals that when \gptj is the reference, the trends in $\tvd$ mimic what we would expect from a metric that interpolates between forward and reverse cross-entropies. Since $\tvd$ is motivated as a metric that balances between precision and recall, our results therefore make intuitive sense. 
On the other hand, the observed trends for the empirical distribution do not have a clear explanation. 

Critically, we find that the observed trends are stable across various design choices; see \cref{app:results} for results with the WikiText dataset and with different choices of $\varepsilon$ for the $\varepsilon$-smoothed versions of metrics.\footnote{We also observed that trends were very stable across the choice of reference model, i.e., using GPT2-XL and the 1.5B parameter version of GPT-Neo rather than \gptj. We omit these results from the appendix to reduce clutter.}\looseness=-1

\begin{figure*}
    \centering
    \includegraphics[width=\linewidth]{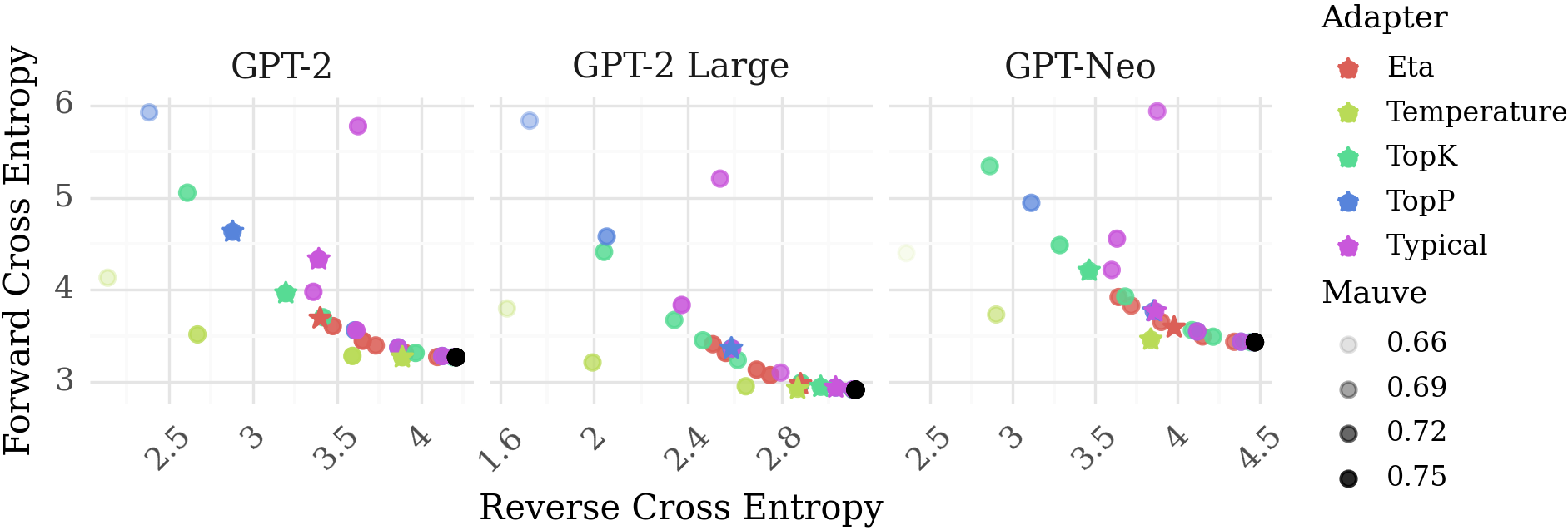}
    \caption{Reverse cross-entropy versus forward cross-entropy (the latter uses $\varepsilon$-smoothing) of the model with GPT-J for various sampling adapter and hyperparameter settings. Stars correspond to values at which hyperparameter settings achieved the highest \mauve scores. 
    The black dot corresponds to ancestral sampling. }
    \label{fig:pr}
\end{figure*}

\paragraph{A Precision--Recall Trade-Off.} 
We next look at whether the shifts induced by common sampling adapters correspond to a precision--recall trade-off according to our probabilistic measures. In \cref{fig:pr}, we compare the reverse and forward cross-entropies (with GPT-J used as the reference) across the adapter hyperparameter settings used. Results using the empirical distribution are similar (see \cref{fig:pr_empirical} in \cref{app:results}). 
\cref{fig:pr} indeed suggests a quite direct trade-off between our operationalizations of precision and recall.  Notably, the highest sequence-level quality scores do not correspond with the sampling adapter--hyperparameter settings that achieve the best precision (i.e., lowest reverse cross-entropy).\footnote{\mauve scores for all adapter--hyperparameter settings and both datasets can be seen in \cref{fig:mauve}.} Rather, they correspond to an intermediate point along the line, suggesting the importance of balancing precision and recall.\looseness=-1

\newcommand{\minusspace}{\phantom{-}}

\paragraph{Correlations.} 
The previous observations motivate us to look at correlations between (sub)word-level probabilistic measures and sequence-level quality metrics. We consider both the WebText and WikiText results when computing correlations. 
In \cref{tab:corr_table}, we see that the reverse $\KL$ of the generation model with GPT-J has the highest (rank) correlation with our quality metrics, closely followed by \tvd. This finding suggests that reverse $\KL$ with another model could be a useful metric for selecting sampling adapter's hyperparameters, as its computation is much faster than standard methods for choosing such hyperparameters---e.g., human annotations or sequence-level quality scores---which require the generation of full sequences.

\section{Related Work}\label{sec:related}

\paragraph{Precision and Recall in Language Generation.}
This is by no means the first work to focus on the notions of precision and recall in the context of language generation. 
Language generator evaluation metrics have historically intentionally prioritized precision-based measures due to their higher correlation with human quality judgments.  
For example, \textsc{bleu} \cite{papineni-etal-2002-bleu} is computed using $n$-gram precision, and the original work on \textsc{chrF} \cite{popovic-2015-chrf}, which is a precision--recall-based metric, found that variants of the metric that placed more weight on precision correlated better with human judgments.  
More recently, \citet{pimentel2023on} report that the reverse $\KL$ divergence between multinomial distributions over embeddings of text from language models and of text from humans correlated more with human quality judgments than the results of other divergence measures. 
On the other hand, measures that place higher importance on recall of the model with respect to some test set, such as perplexity, are known not to be good indicators of text quality \cite{holtzman_curious_2020,pmlr-v97-cohen19a,meister+al.pre22}.
In terms of model training, alternative objectives that emphasize precision have been proposed in an attempt to alleviate the zero-avoiding effect induced by optimization for maximum likelihood~\cite{kang-hashimoto-2020-improved, pang2021text}.\looseness=-1

\begin{table}
    \centering
    \small
    \adjustbox{max width=\linewidth}{
    \begin{tabular}{ll|cllll}
      & &	& \multicolumn{2}{c}{$\KL$} & \multicolumn{2}{c}{\textbf{Cross-entropy}} \\ \cmidrule(lr){4-5}\cmidrule(lr){6-7}
      & &	\textbf{TVD}		& \textbf{Reverse}	&	\textbf{$\varepsilon$-Forward} & \textbf{Reverse}	&	\textbf{$\varepsilon$-Forward} \\
       \toprule
\multirow{3}{*}{\rotatebox[origin=c]{90}{\textbf{GPT-J}}} & GPT-2 &\,\,-0.73\threeS	&-0.77\threeS	&		-0.38\threeS & -0.11 &-0.44\threeS\\
& GPT-Neo &\,\,	-0.74\threeS	&-0.73\threeS	&	-0.33\threeS & \minusspace0.08	 &-0.41\threeS\\
& GPT-Large &\,\,	-0.77\threeS	&-0.80\threeS	&	-0.49\threeS & \minusspace0.01 & -0.55\threeS \\
\midrule
\multirow{3}{*}{\rotatebox[origin=c]{90}{\textbf{Empirical}}} & GPT-2	&\,\,-0.18\threeS	&-0.26\threeS	&	-0.48\threeS & -0.18\threeS&-0.48\threeS\\
& GPT-Neo &\,\,-0.02	&-0.25\threeS	&		-0.42\threeS &-0.02&-0.42\threeS\\
& GPT-Large &\,\,-0.10	&-0.50\threeS 	&		-0.61\threeS &-0.10 &-0.61\threeS\\

    \end{tabular}}
    \vspace{3pt}
    \caption{Spearman correlations of (sub)word-level probabilistic measures with \mauve. We use  \threeS\, to indicate significance with a $p$-value $<0.001$. }
    \label{tab:corr_table}
\end{table}

\paragraph{Analysis of Language Generation Models.}

The effect of sampling adapters on language models has previously been discussed in the framework of a quality--diversity trade-off~\cite{zhang_trading_2020, meister-etal-2022-high}. 
For instance, \citet{nadeem-2020-systematic} and \citet{wiher+al.tacl22} catalog various sampling adapters and analyze their properties with respect to a quality--diversity trade-off using a wide range of automatic metrics. \citet{hashimoto-etal-2019-unifying} propose an evaluation framework that combines human and statistical evaluation. In contrast, our work makes an explicit connection to the concepts of precision and recall and analyzes the effect of sampling adapters employing measures of differences in distributions. While \citet{pillutla2021mauve} likewise use notions of precision and recall for assessing language generators, they look at quantized distributions over language embedding spaces rather than directly at distributions over (sub)words.\looseness=-1

\section{Conclusion}
In this work, we offer a formal treatment of sampling adapters and provide an analysis that aims to uncover why they are effective when used with probabilistic models for language generation.  
To this end, we first introduce a general framework that encompasses most of the transformations performed by previously proposed sampling adapters. 
We then offer an intuition as to why sampling adapters may lead to better language generators.
Using the notions of precision and recall proposed for generative models, which can be quantified in terms of standard probabilistic measures, we perform an empirical analysis.
We find evidence that the application of sampling adapters increases the precision of a distribution at the expense of its recall; this observation is robust across several experimental design choices. %
We further find a high correlation between sequence-level quality metrics and reverse $\KL$ divergence of the generation model with a reference model.\looseness=-1 
\section*{Acknowledgments}
We would like to thank John Hewitt and Afra Amini for the insightful discussions preceding this work. 
Clara was supported by a Google Ph.D. Fellowship. 
Tiago was supported by a Facebook Ph.D. Fellowship. 
Ethan was supported by an ETH Zürich Postdoctoral Fellowship. 

\section*{Limitations}
A clear limitation of this work is that the results have been shown only for English. Further work should consider other model architectures, as well as datasets that span a variety of languages and domains. Another limitation is that we do not conduct human evaluations. Given the large number of adapter and hyperparameter settings that we chose to explore, acquiring the human evaluations that would have allowed us to make statistically significant conclusions regarding the relationships between text quality, distribution-level measures, and adapter--hyperparameter settings would have been financially prohibitive. Instead, we chose to look at automatic sequence-level quality metrics that are known to correlate highly with human quality judgments. Further, it has been observed that crowd-sourced judgments of text quality are far from perfect \cite{clark2021gold}, making it not obvious whether this is indeed the better option.

\section*{Ethical Considerations}
The use of language models for text generation comes with several ethical concerns. 
Especially when using sampling-based decoding algorithms, as is promoted in this work, the text generated by probabilistic models may contain malicious or hallucinatory content. 
This may be an intention of the user, but can also occur simply due to the training data that the model was exposed to, which is often not carefully filtered for undesirable material that a model then learns to mimic. 
The goal of works like this---to help create systems that can produce more human-like text---may also make it easier to automatically produce such content, which can ultimately have several negative downstream side effects. 
We caution designers and users of text generation systems to publicly advertise when content was created by a machine, and implement checks to prevent the production of harmful material.

\bibliography{anthology,custom}
\bibliographystyle{acl_natbib}
\newpage
\clearpage

\appendix
\onecolumn

\section{Additional Results}\label{app:results}
\begin{figure*}[!htb]
    \centering
\begin{subfigure}[b]{\textwidth}
   \includegraphics[width=\linewidth]{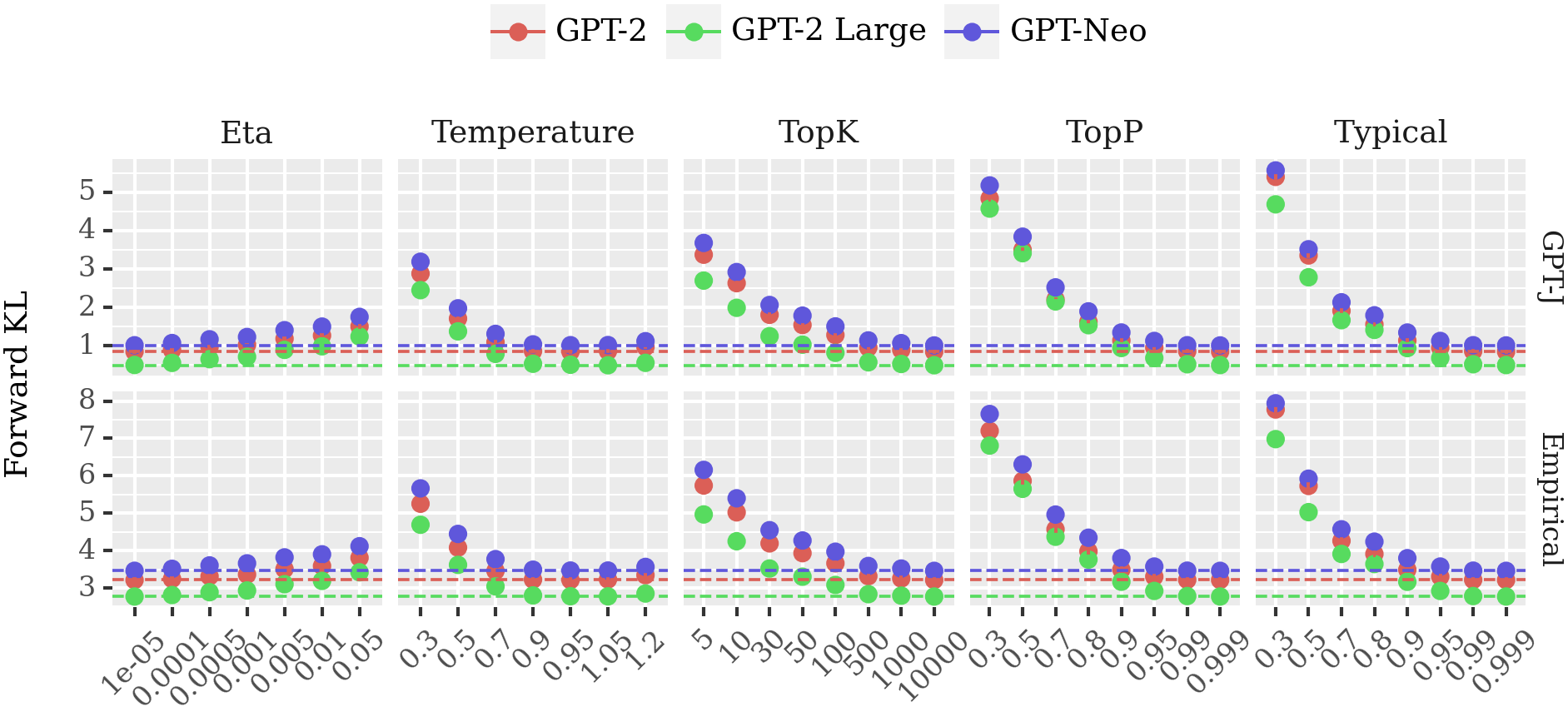}
\end{subfigure}
    \begin{subfigure}[b]{\textwidth}
   \includegraphics[width=\linewidth]{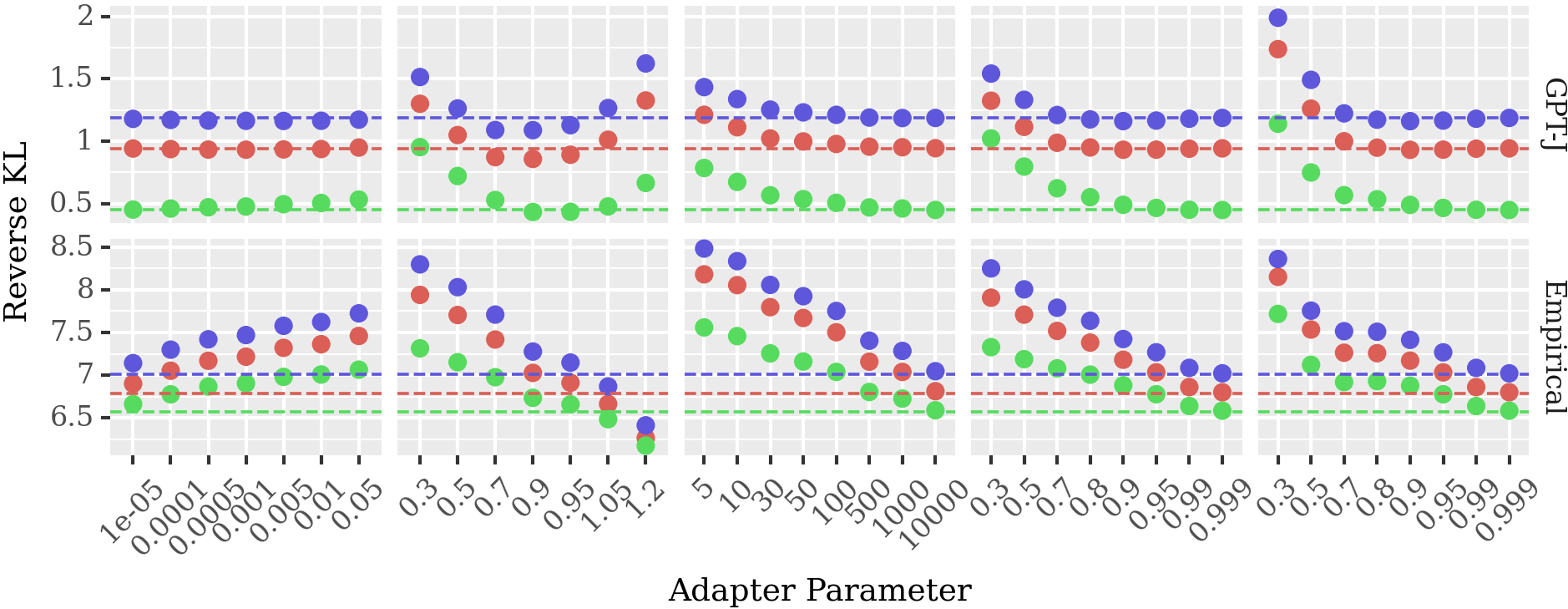}
\end{subfigure}

    \caption{Reverse and forward $\KL$ divergence of the model with GPT-J and the empirical distribution (WebText test set) after different sampling adapter methods have been applied to the output distribution. Note that the $\varepsilon$-method, as described in \cref{sec:pr}, is used in all but reverse $\KL$ estimates of models with GPT-J. Dashed lines represent divergence with unmodified distribution, i.e., the equivalent of using ancestral sampling. }\label{fig:backward_forward_kl}
\end{figure*}
\begin{figure*}[!ht]
    \centering
    \includegraphics[width=\linewidth]{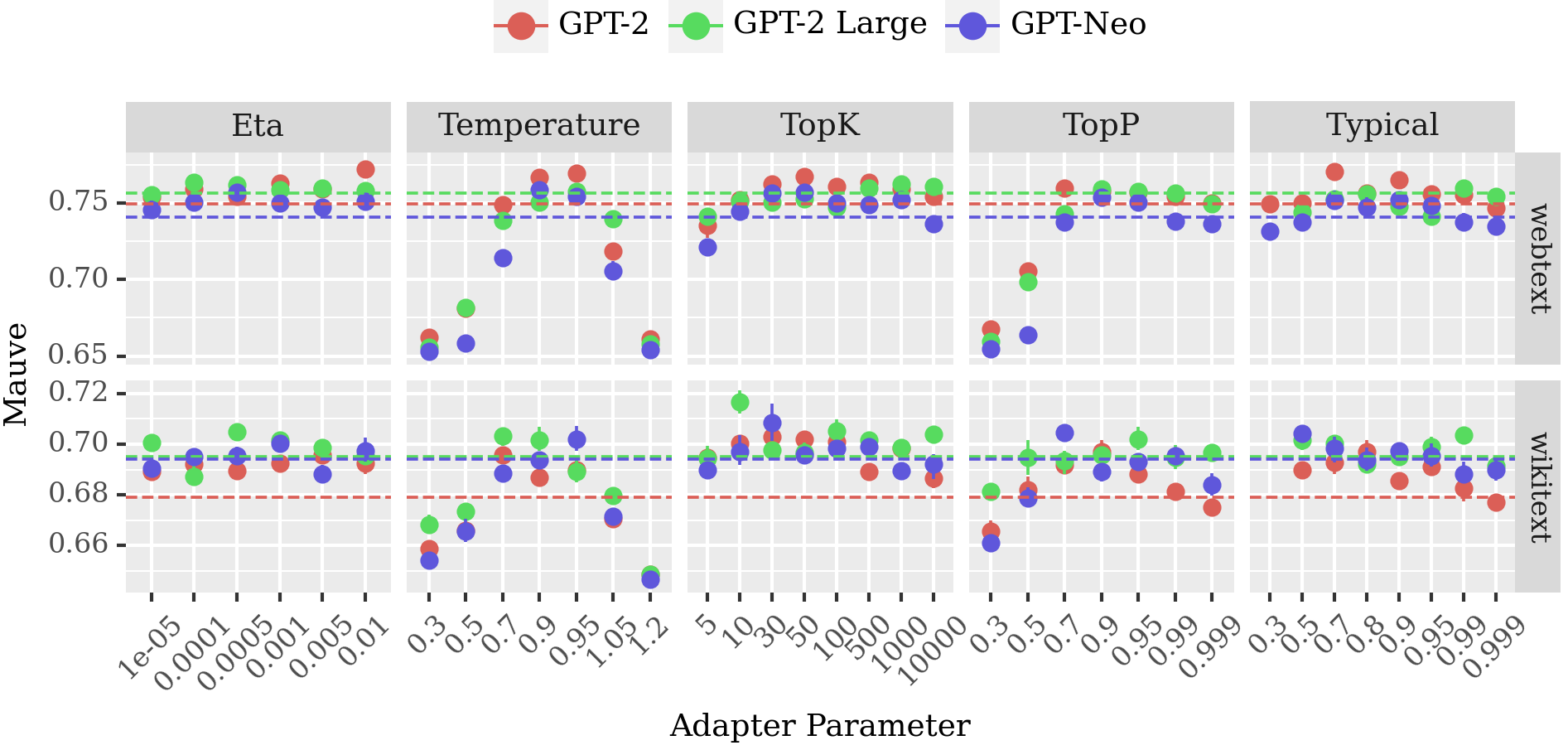}
    \caption{\mauve scores for text generated using WebText prefixes and different sampling adapters. The dashed lines indicate the scores of samples generated using ancestral sampling.}
    \label{fig:mauve}
\end{figure*}
\begin{figure*}
    \centering
    \includegraphics[width=\linewidth]{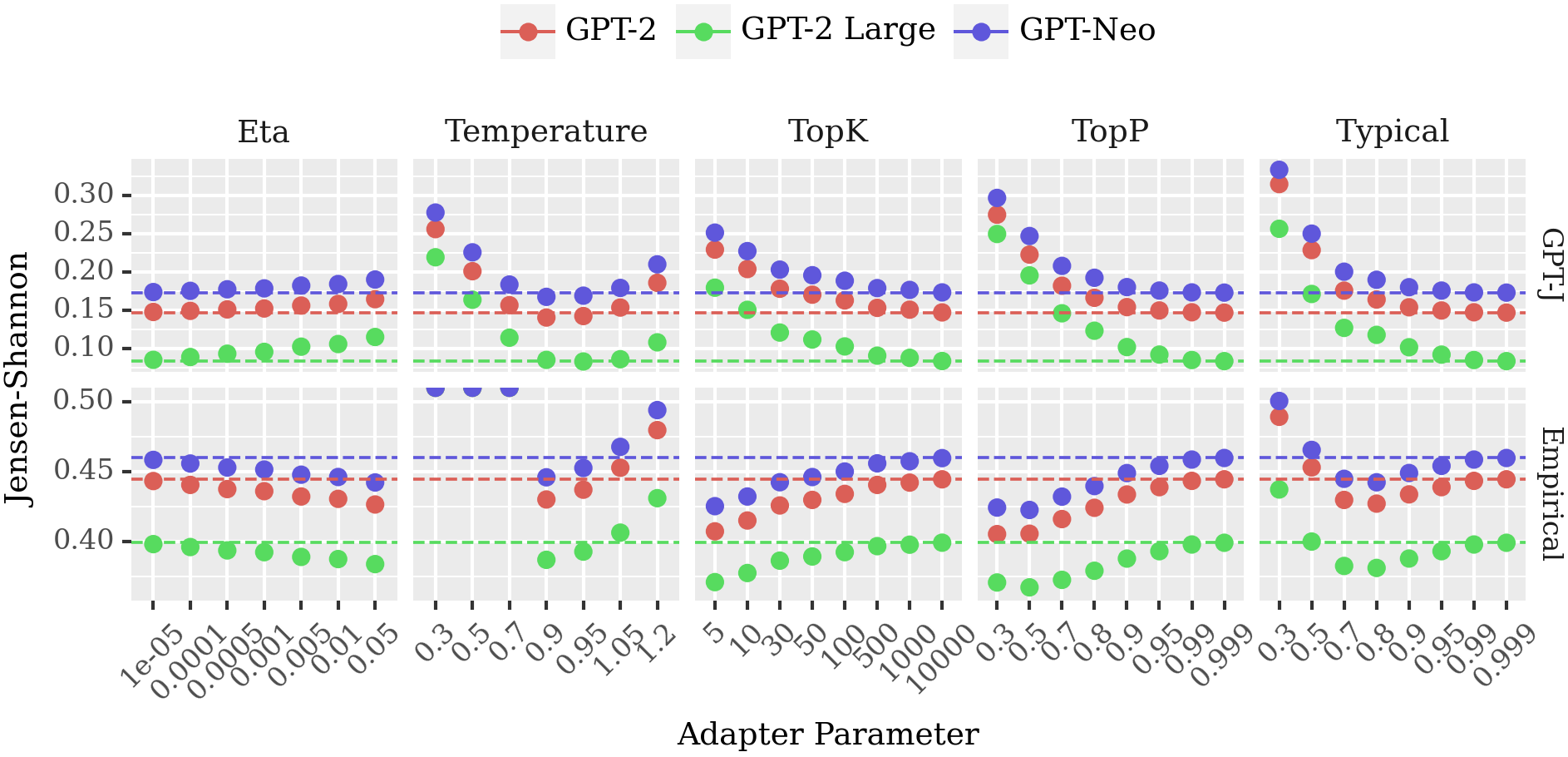}
    \caption{JS divergence of the model with the empirical distribution in the first row and with GPT-J in the second row after different sampling adapter methods have been applied to the output distribution. Dashed lines represent the distance to the unmodified distribution. We observe that at lower temperature values, some NaNs are produced by the $\JS$ computation with the empirical distribution.}
    \label{fig:js}
\end{figure*}

\begin{figure*}
    \centering
    \includegraphics[width=\linewidth]{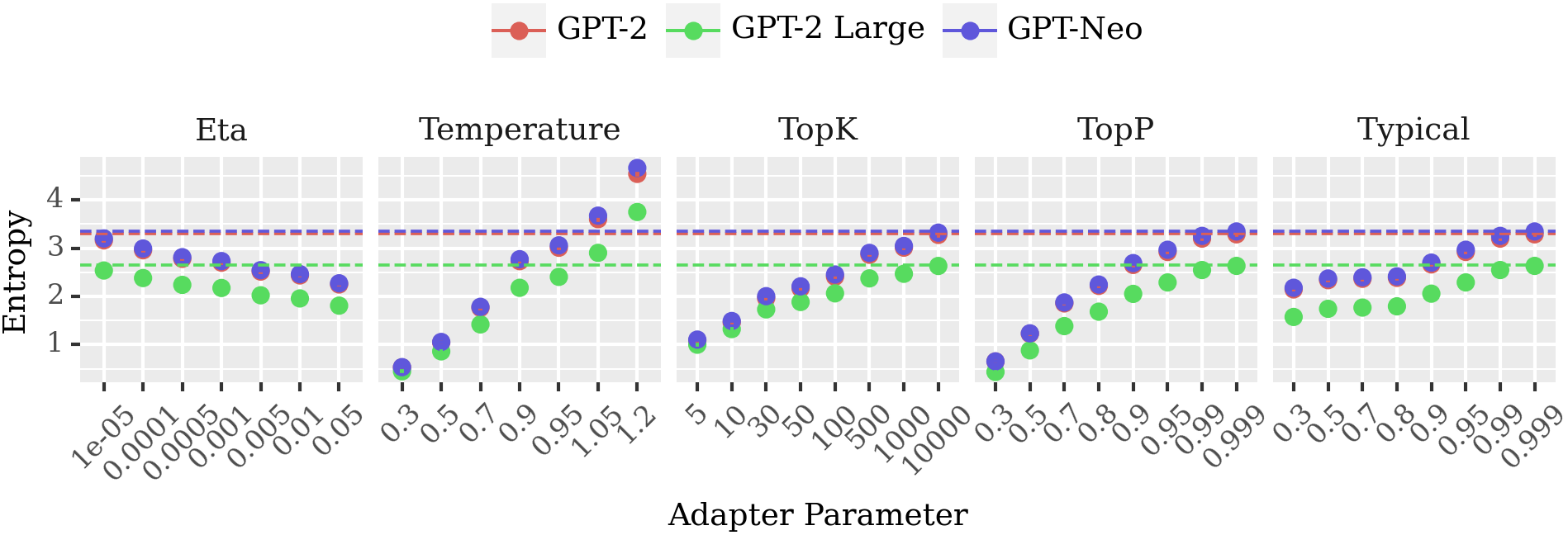}
    \caption{ Average entropy of the distribution $\adapterdist(\cdot\mid \yy_{<t})$ for different sampling adapter--hyperparameter combinations. Dashed lines correspond to the entropy of the unmodified distribution. }
    \label{fig:token}
\end{figure*}
\begin{figure*}
    \centering
    \includegraphics[width=\linewidth]{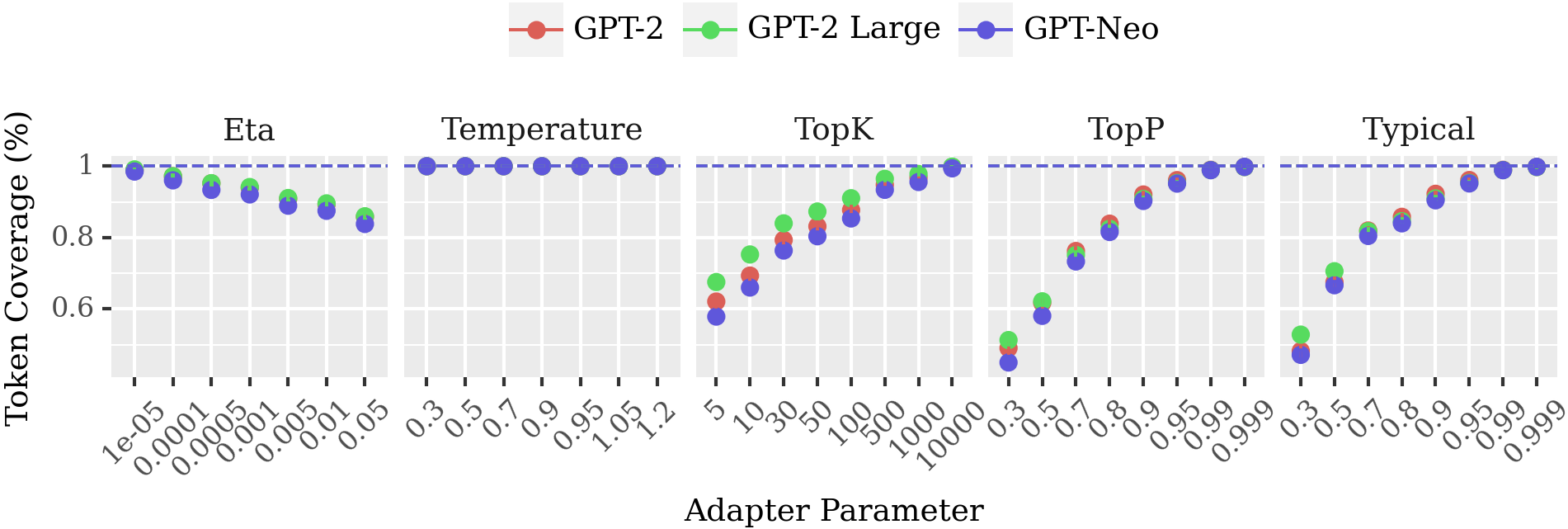}
    \caption{Average model token coverage \emph{per sequence $\yy$} (i.e., percentage of tokens to which the adapter assigns non-zero probability) of the WebText test set after different sampling adapter methods have been applied to the output distribution. Dashed lines correspond to unmodified distribution, which always assigns probability mass to each token. }
    \label{fig:ent}
\end{figure*}

\begin{figure*}[!htb]
    \centering
   \includegraphics[width=\linewidth]{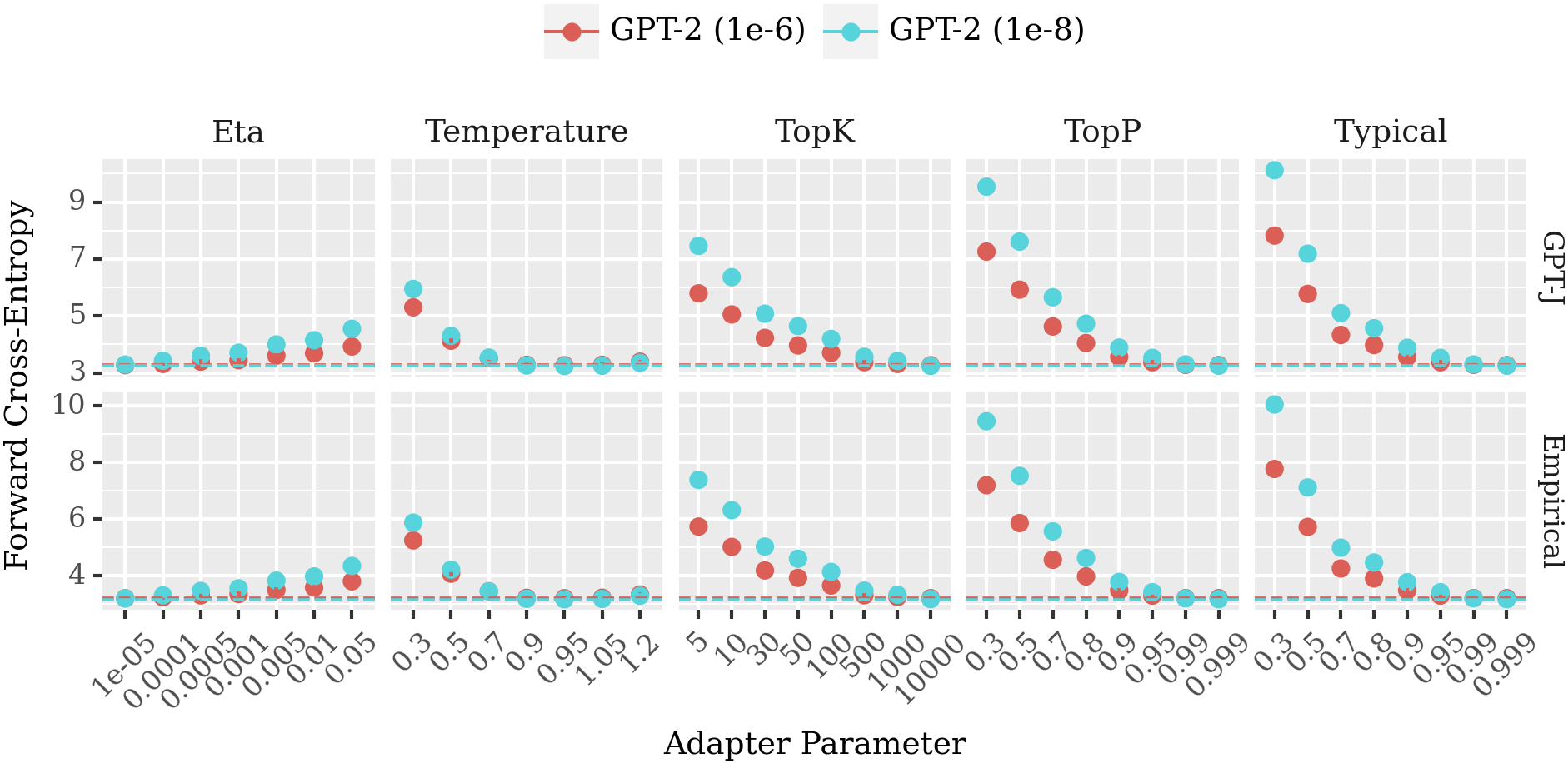}
    \caption{Same plot as \cref{fig:backward_forward} albeit using smaller $\varepsilon$ (1e-8 instead of 1e-6) in computation of $\epsilon$ variants of methods. Results are essentially unchanged, except for a slight shift in axis values.}
\end{figure*}

\begin{figure*}[!htb]
    \centering
\begin{subfigure}[b]{\textwidth}
   \includegraphics[width=\linewidth]{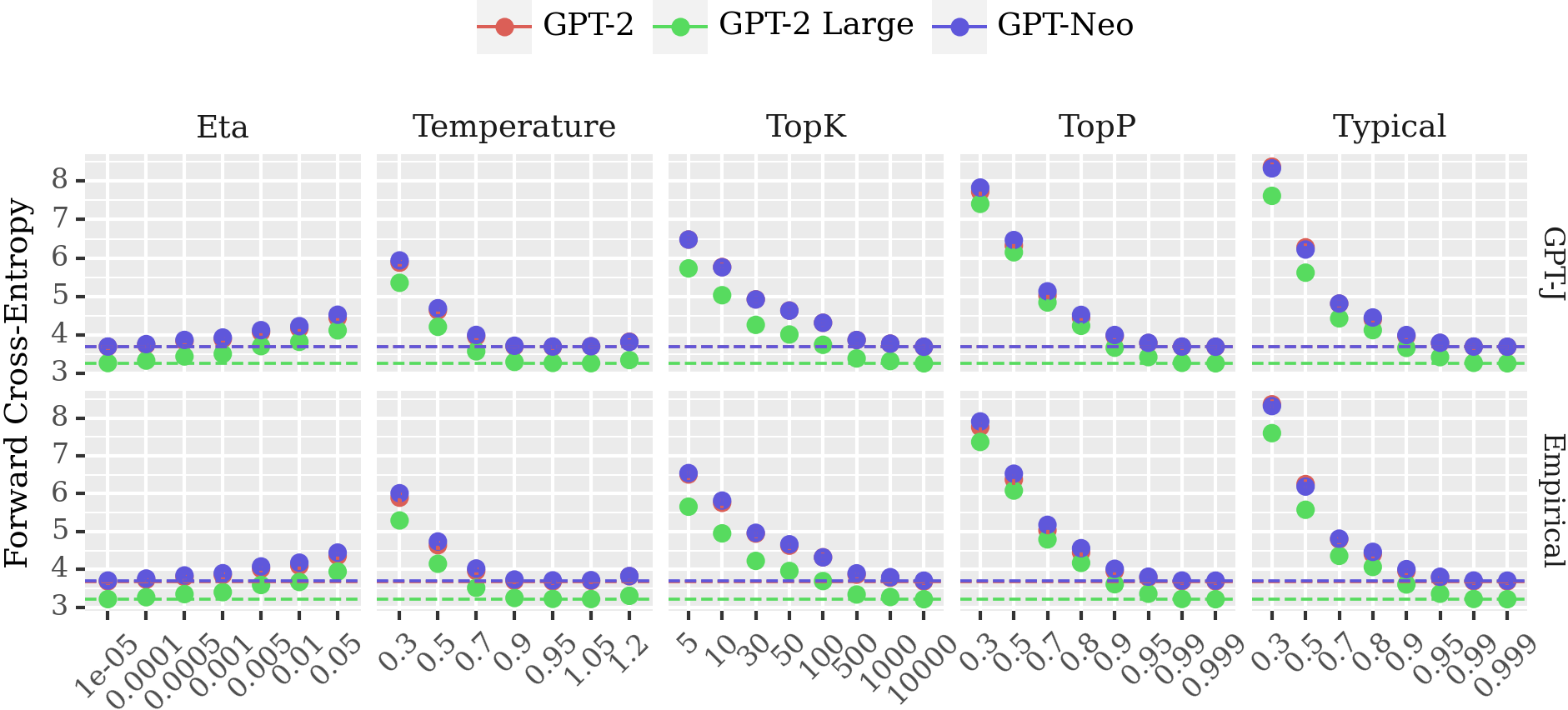}
\end{subfigure}
\begin{subfigure}[b]{\textwidth}
   \includegraphics[width=\linewidth]{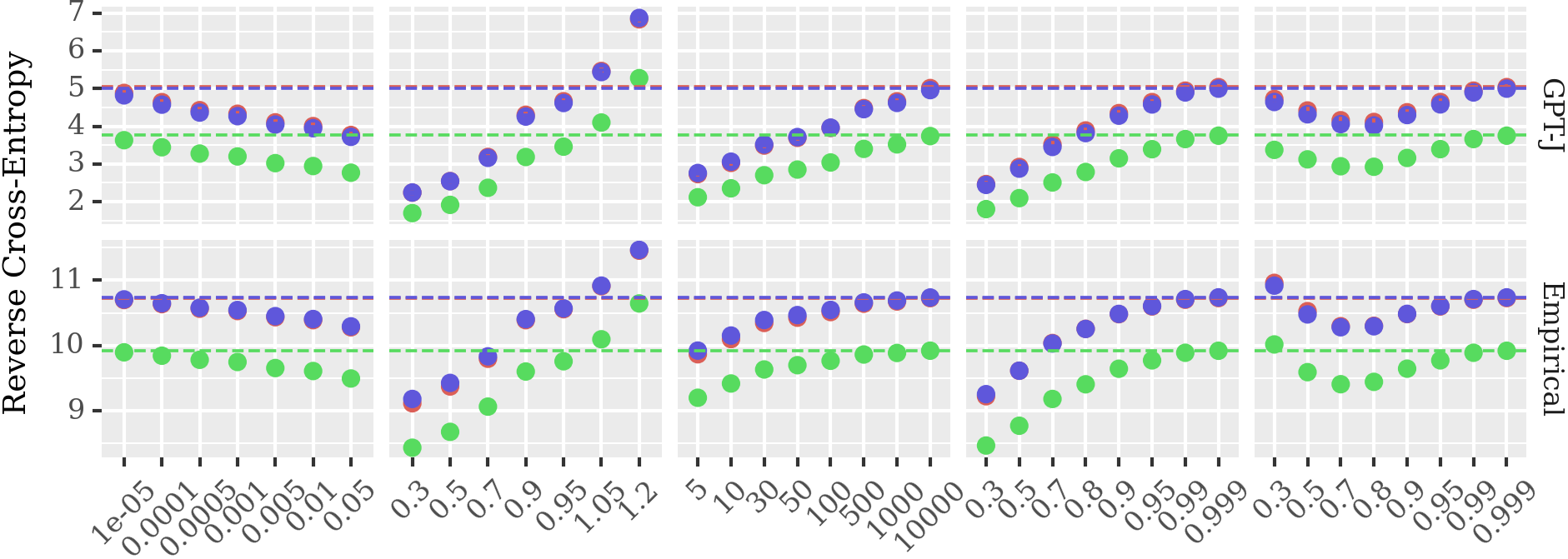}
\end{subfigure}
\begin{subfigure}[b]{\textwidth}
   \includegraphics[width=\linewidth]{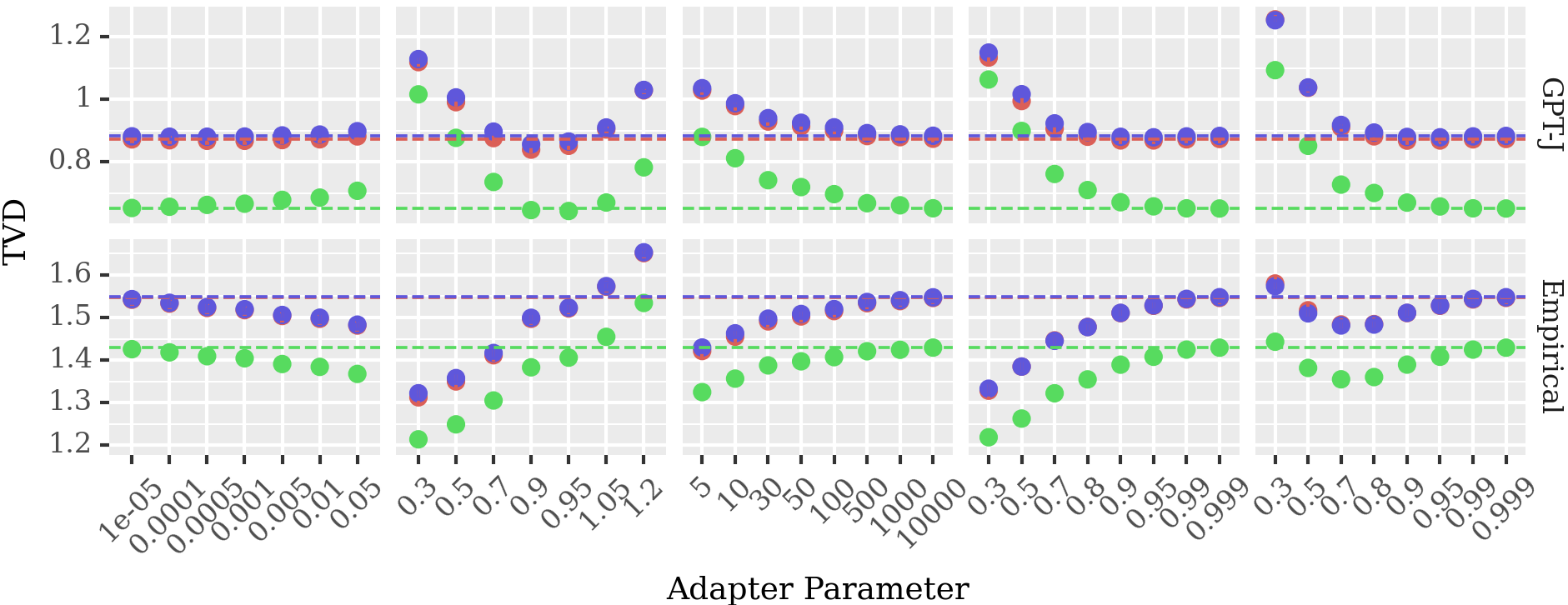}
\end{subfigure}
    \caption{Same plot as \cref{fig:backward_forward} except using the test set of WikiText as our set of strings ($\yy$) and to construct the empirical distribution.}
\end{figure*}

\begin{figure*}
    \centering
    \includegraphics[width=\linewidth]{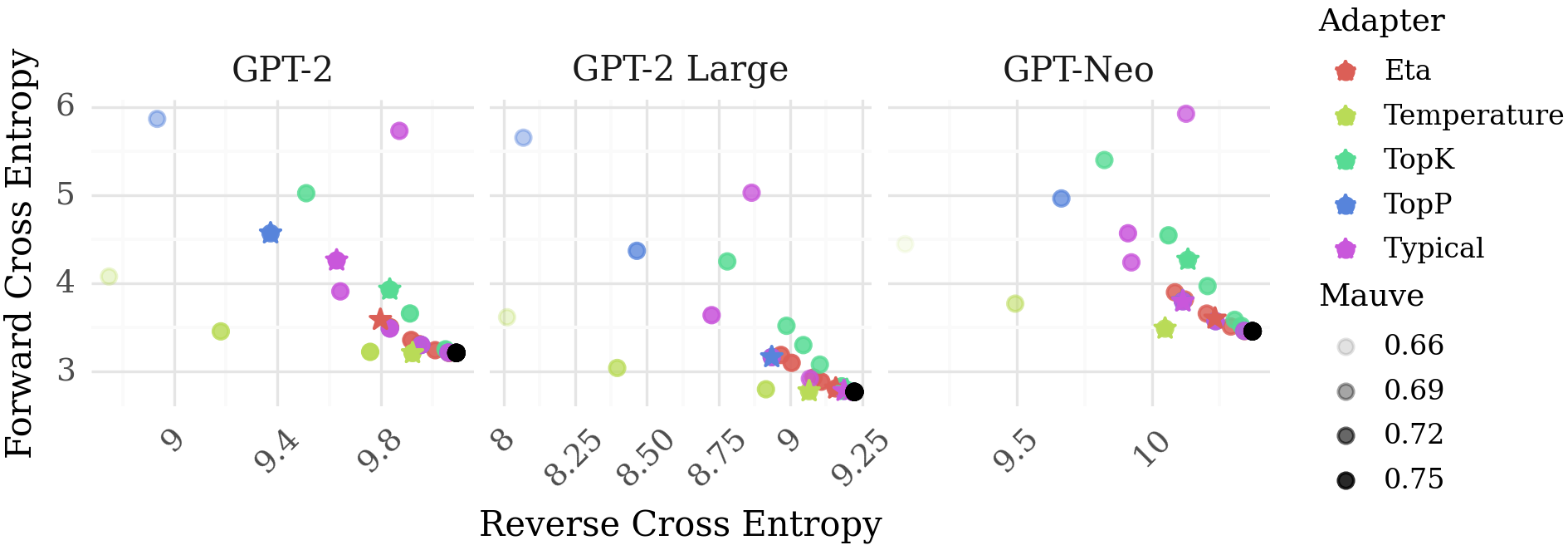}
    \caption{Reverse cross-entropy versus forward cross-entropy divergence (both using $\varepsilon$-smoothing) of the model with the empirical distribution for various sampling adapter and hyperparameter settings. Stars correspond to values at which hyperparameter settings achieved the highest \mauve scores. The black dot corresponds to ancestral sampling. }
    \label{fig:pr_empirical}
\end{figure*}

\end{document}